\documentclass[journal]{template/vgtc}                

\usepackage{pifont}

\usepackage{xcolor}

\usepackage{xspace}
\newcommand{\ie}{i.e.,\xspace}

\newcommand{\eg}{e.g.,\xspace}

\usepackage{mwe}
\usepackage{enumitem}

\newcommand\myparagraph[1]{\textbf{#1}.}

\newcommand{\tool}{\textsc{VisQA}}
\newcommand{\toolurl}{{\small \url{https://visqa.liris.cnrs.fr}}}
\newcommand{\giturl}{{\small \url{https://github.com/Theo-Jaunet/VisQA}} }

\newif\ifisreview

\newcommand{\cond}[1]{\textcolor[rgb]{0, 0, 0}{#1}}

\newcommand{\added}[1]{\textcolor[rgb]{0, 0, 0}{#1}}

\usepackage{bm}
\usepackage{amssymb,amsmath}

\usepackage{pifont}

\ifpdf
  \pdfoutput=1\relax                   
  \usepackage{graphicx}                
  \DeclareGraphicsExtensions{.pdf,.png,.jpg,.jpeg} 
\else
  \usepackage{graphicx}                
  \DeclareGraphicsExtensions{.eps}     
\fi%

\graphicspath{{figures/}{pictures/}{images/}{./}} 

\usepackage{microtype}                 
\PassOptionsToPackage{warn}{textcomp}  
\usepackage{textcomp}                  
\usepackage{mathptmx}                  
\usepackage{times}                     
\usepackage{cite}                      
\usepackage{tabu}                      
\usepackage{booktabs}                  

\onlineid{0}

\vgtccategory{Research}
\vgtcpapertype{please specify}

\title{VisQA: X-raying Vision and Language Reasoning in Transformers}

\author{Th\'eo~Jaunet*,
        Corentin Kervadec*,
        Romain~Vuillemot,
        Grigory Antipov,
        Moez Baccouche,
        and~Christian~Wolf}
\authorfooter{
\item (*) indicates equal contribution
    \item T. Jaunet  and C. Wolf are with INSA-Lyon and LIRIS.
    \item C. Kervadec is with INSA-Lyon, LIRIS, and Orange Innovation.\\%
    \item R. Vuillemot is with \'Ecole Centrale de Lyon and LIRIS.\\%
    \item G. Antipov and M. Baccouche are with Orange Innovation.
}

\usepackage{xcolor}

\usepackage{xspace}

\usepackage{enumitem}

\shortauthortitle{Biv \MakeLowercase{\textit{et al.}}: Global Illumination for Fun and Profit}
 
\abstract{
Visual Question Answering systems target answering open-ended textual questions given input images. They are a testbed for learning high-level reasoning with a primary use in HCI, for instance assistance for the visually impaired. Recent research has shown that state-of-the-art models tend to produce answers exploiting biases and shortcuts in the training data, and sometimes do not even look at the input image, instead of performing the required reasoning steps. We present VisQA, a visual analytics tool that
explores this question of reasoning vs. bias exploitation. It exposes the key element of state-of-the-art neural models --- attention maps in transformers. Our working hypothesis is that reasoning steps leading to model predictions are observable from attention distributions, which are particularly useful for visualization. The design process of VisQA was motivated by well-known bias examples from the fields of deep learning and vision-language reasoning and evaluated in two ways. First, as a result of a collaboration of three fields, machine learning, vision and language reasoning, and data analytics, \cond{the work lead to a better understanding of bias exploitation of neural models for VQA, which eventually resulted in an impact on its design and training through the proposition of a method for the transfer of reasoning patterns from an oracle model}. Second, we also report on the design of VisQA, and a goal-oriented evaluation of VisQA targeting the analysis of a model decision process from multiple experts, providing evidence that it makes the inner workings of models accessible to users.
} 


\CCScatlist{ 
 \CCScat{K.6.1}{Management of Computing and Information Systems}%
{Project and People Management}{Life Cycle};
 \CCScat{K.7.m}{The Computing Profession}{Miscellaneous}{Ethics}
}

\teaser{
 \includegraphics[width=0.95\linewidth]{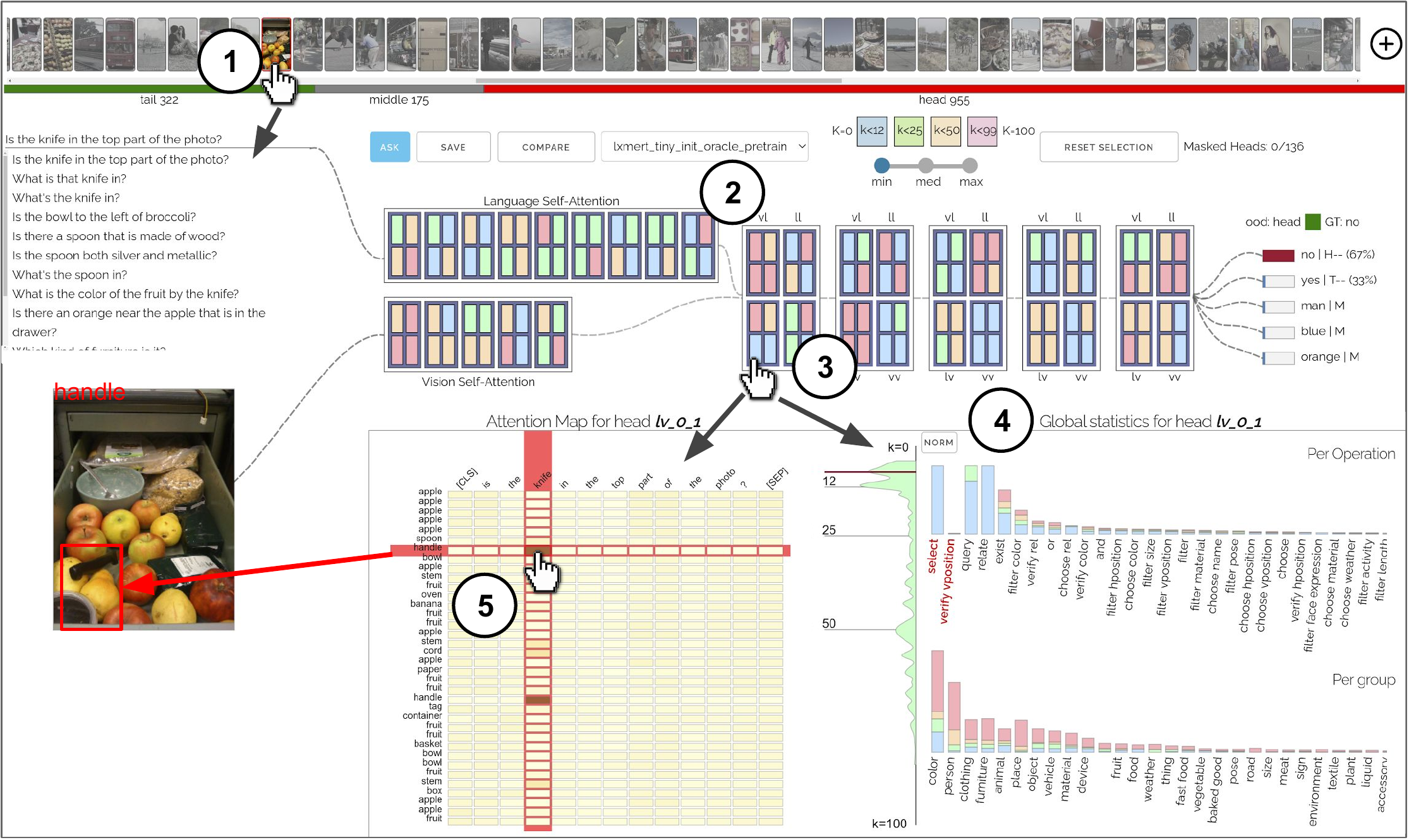}
 \centering
  \caption{Opening the black box of neural models for vision and language reasoning: given an open-ended question and an image~\ding{192}, \tool\ enables to investigate whether a trained model resorts to reasoning or to bias exploitation to provide its answer. This can be achieved by exploring the behavior of a set of attention heads~\ding{193}, each producing an attention map~\ding{196}, which manage how different items of the problem relate to each other. Heads can be selected~\ding{194}, for instance, based on color-coded activity statistics. Their semantics can be linked to language functions derived from dataset-level statistics~\ding{195}, filtered and compared between different models.}
\label{fig:teaser}
}

\vgtcinsertpkg

\begin{document}


\firstsection{Introduction}

\maketitle
 
Visual Question Answering (VQA) systems~\cite{antol2015vqa} attempt to answer questions provided as input in a textual form together with a corresponding image. As an example, asking the question \emph{``Is the knife in the top part of the photo?''} associated with the input image shown in Figure~\ref{fig:teaser} should yield the answer \emph{``No''}.
Direct applications of such systems are support for the visually impaired, semi-autonomous robot navigation through language instructions, and, more generally, AI tools covering a broad spectrum of tasks guided through language input. In particular, VQA serves as a testbed for learning high-level reasoning from data, as the performance of targeted models and methods relies on advances in Computer Vision (CV), Natural Language Processing (NLP), and Machine Learning (ML). The task deals with large varieties, and solving an instance can involve visual recognition, logic, arithmetic, spatial reasoning, intuitive physics, causality, and multi-hop reasoning. It also requires combining two modalities of different nature: images and language.

Recent VQA models are based on a powerful type of deep neural network called transformers~\cite{vaswani2017attention}. Originally developed for NLP tasks, these models have been extensively applied to VQA~\cite{Yu_2019_CVPR,Tan2019LXMERTLC,lu2019vilbert} and recently even on pixel-level in pure image-based problems ~\cite{NLNLCVPR2018,GirdharCVPR2020,ZhaoJiaKolutionCVPR2020,dosovitskiy2021an,SSTVOSCVPR2021}. Transformers are conceptually simple models, which, however, can learn very complex relationships between the items of unordered sets, each of which is represented as a (learned) embedding in a vector space. Making sense of a learned neural model and verifying its inner workings is a difficult problem, which we address in this work.

In this paper, we focus on a typical and important problem arising with trained neural models, and in particular models for vision and language reasoning: as they are trained with supervision to provide correct answers, they often tend to find shortcuts in learning and learn to exploit spurious biases in training data instead of the desired reasoning a human would apply in a similar situation~\cite{agrawal2016analyzing,goyal2017making,kervadec2020roses, manjunatha2019explicit}. To provide an example, if a model is asked \emph{``What is the color of the banana''?} with an image of a \emph{``banana''}, it might learn to answer \emph{``yellow''} whatever the real color of the image is, as this is probably the correct answer for a large majority of input images --- solving the correct reasoning problem is a much harder task. An exact definition of the term ``correct reasoning'' is difficult, we refer to ~\cite{bottou2014machine,kervadec2020roses} and define it as \emph{algebraically manipulating words and visual objects to answer a new question}. In particular, we interpret reasoning as the opposite of exploiting spurious biases in training data.

Existing work on bias reduction and the evaluation of bias origins tends to focus on statistical techniques, whose power lies in quantitative evaluation and visualizations on dataset-level, showing full or marginal distributions of inputs, features, and outputs, and resort to dimensionality reduction. While these techniques are very useful, their power is limited when we search for insights into detailed inner workings of neural models, for which an investigation per sample is more helpful. Only for a single instance, it is possible to observe the origins for lack of reasoning, which can include, aside from errors in the trained reasoning module itself, also problems in the input pipeline (the object detection module) and wrong annotations of ground truth data.

We introduce \tool, an instance-based visual analytics tool designed to help domain experts, referred to as model builders~\cite{Hohman2019VisualFrontiers}, investigate how information flows in a neural model and how the model relates different items of interest to each other in vision and language reasoning. Attention maps are at the heart of transformer-based deep networks, and as such are the primary object studied by \tool. It allows an expert to browse through image and question pairs sorted by an automatic estimate of the amount of reasoning that went into answering each sample. Once a pair is selected, users can explore the different attention maps represented as heatmaps. The exploration is guided by their position in the model, but also by color codes that convey the intensity of each head, i.e. whether they focus attention narrowly on specific items, or broadly over the full input set. Complementary dataset-wide statistics are provided for each selected attention head, either globally, or with respect to specific reasoning modes of language functions, e.g. \emph{``What is'', ``Where is``, ``What color''} etc.
While the tool is post-hoc, it is also interactive and allows certain modifications to the internal structure of the model. At any time, attention maps can be pruned to observe their impact on the output answer.

\tool\ is the result of a collaboration between experts in visual analytics, and experts in Visual Question Answering systems and Machine Learning. 
As will be detailed in section~\ref{sec:use-case}, this collaboration, and data gathered using \tool, led to improvements of the reasoning capabilities of transformer-based models by introducing new methodological contributions in machine learning and computer vision, which we also recently reported in a different associated publication~\cite{uscvpr2021}. The usability of \tool\ has been evaluated by different experts in deep learning, who were not involved in the project nor its design. We report experiments with qualitative interviews and results in section~\ref{sec:eval}.

 
\added{In this work, we contribute to a better understanding of bias in VQA models as follows:}

\begin{itemize}

    \item \added{\textbf{\tool\ an interactive visual analytics tool} which helps experts to explore the inner workings of transformers models for VQA by displaying models' attention heads in an instance-based fashion.}
    
    
    \item \added{\textbf{A set of visualizations to address bias in VQA systems} designed to explore models' performances in real-time with altered attention, and/or by asking free-text questions.}
    
    \item \added{\textbf{Insights on the emergence bias in transformers for VQA} gained by experts through an in-depth analysis using \tool, along with an evaluation of its usability to estimate models' predictions and eventually bias exploitation.}
\end{itemize}

\added{
\tool\ is available online as an interactive prototype~\toolurl, and our code and data are available as an open-source project: ~\giturl.
}

\section{Background}

We first introduce some background on understanding neural networks in vision and language reasoning, the context of this paper, and we will provide a short and concise introduction into transformers, the type of neural networks which currently dominates academic and industrial research in language reasoning, and in vision-based language problems.

\subsection{Transformers and Attention} 
\label{sec:bgtransformers}
Following the introduction and success of transformers applied to natural language processing tasks~\cite{vaswani2017attention,devlin2019bert}, transformer-based models were also proposed for VQA~\cite{gao2019dynamic, Yu_2019_CVPR}.
Their key strength is the ability to contextualize input representations, i.e. to take input items like words and objects, each one represented in a vectorial form called ``\emph{embedding}'', and to enrich them, adding information on relationships. This is achieved by series of transformations of the input vectors, which effectively encodes the reasoning process, and the underlying key mechanism is attention (self-attention).
We start with a brief overview of how a typical language transformer works by applying it to encode the question \textit{"What is the name of the clothing item that is white?"}.

\myparagraph{\added{Step 1: Preparation: Sentence Tokenization}}
\added{
We split the question into elementary language items (called ``tokens'') with the WordPiece tokenizer~\cite{wu2016google}.
Two special tokens are added at the beginning and at the end of the sentence (respectively} \textsc{`cls'} and \textsc{`sep'}).
\added{
While the latter encodes the end of the sentence, the former is of the uttermost importance; it is transformed by the model as are the other tokens, with the difference that the} \textsc{`cls'} \added{ token is transformed to encode the task-specific information, and the answer. In the given example, at the end of the transformation, it is expected to contain the information required to predict the name of the white clothing.} 
\added{
Each token (including special tokens) is then projected into a high-dimensional vector space through a learned dictionary, resulting in a sequence of $N$ token embeddings:} $\bm{L}=[\bm{l}_{CLS}, \bm{l}_1, \dots, \bm{l}_i,\dots,\bm{l}_{N-2}, \bm{l}_{SEP}], \bm{l}_i\in\mathbb{R}^n$, $n$~\added{being the (chosen) embedding dimension.}

\myparagraph{\added{Step 2: Attention Maps}}
\added{
Transformers progressively contextualize each of the $N$ input embeddings by a sequence of self-attention operations (layers), with the objective of making each token embedding ``aware'' of the neighboring embeddings.
In our example, it might be helpful to combine the information from the embeddings of tokens \emph{`item'}, \emph{`clothing'} and \emph{`white'} into one ``enriched'' embedding describing the referred object, as the three words are semantically related.
More generally, the so-called ``self-attention operations'' (layers) are the key elements of this type of model. They are implemented via the calculation of \emph{attention maps}} $\bm{A}=\{\bm{\alpha}_{ij}\}$,
\added{
which reflect the $N^2$ pairwise interactions between the tokens, each $\bm{\alpha}_{ij}$ being the similarity between token embeddings $i$ and $j$. The similarity function, here, the scaled dot-product, is calculated between a trainable projection of the embedding $i$, called \textit{query}, and a trainable projection of the embedding $j$, called \textit{key}.
The per-token attention energy is normalized into a probability distribution using a row-wise softmax.
In our example, the row vector} $A_{7}=\{\bm{\alpha}_{7j}\}_{j\in \{0,\dots, N-1\}}$ \added{encodes the $N$ interactions between \emph{`clothing'} and the other words.}

\myparagraph{\added{Step 3: Token Updates}}
\added{
Each token embedding is updated as a linear combination of a trained function of the input embeddings, called \textit{values}, weighted by the attention map $\bm{A}$. Hence, the model transforms each token by learning a strategy for looking at specific other words.}

\myparagraph{\added{Multi-headed Attention}}
\added{
As more classical neural networks, Transformers are organized into a sequence of layers. Each of these layers bundles together \emph{multiple attention heads} working in parallel, which allows the model to learn different cooperating strategies. Different heads might learn different syntactic or semantic language functions, as shown in \cite{voita2019analyzing} for language models, and as we will show in the experimental section for vision and language reasoning.
The outputs of the attention heads are combined with standard neural network blocks.} 




\begin{figure}[t!]
     \vspace{-.1cm}
    \centering
\includegraphics[width=\linewidth]{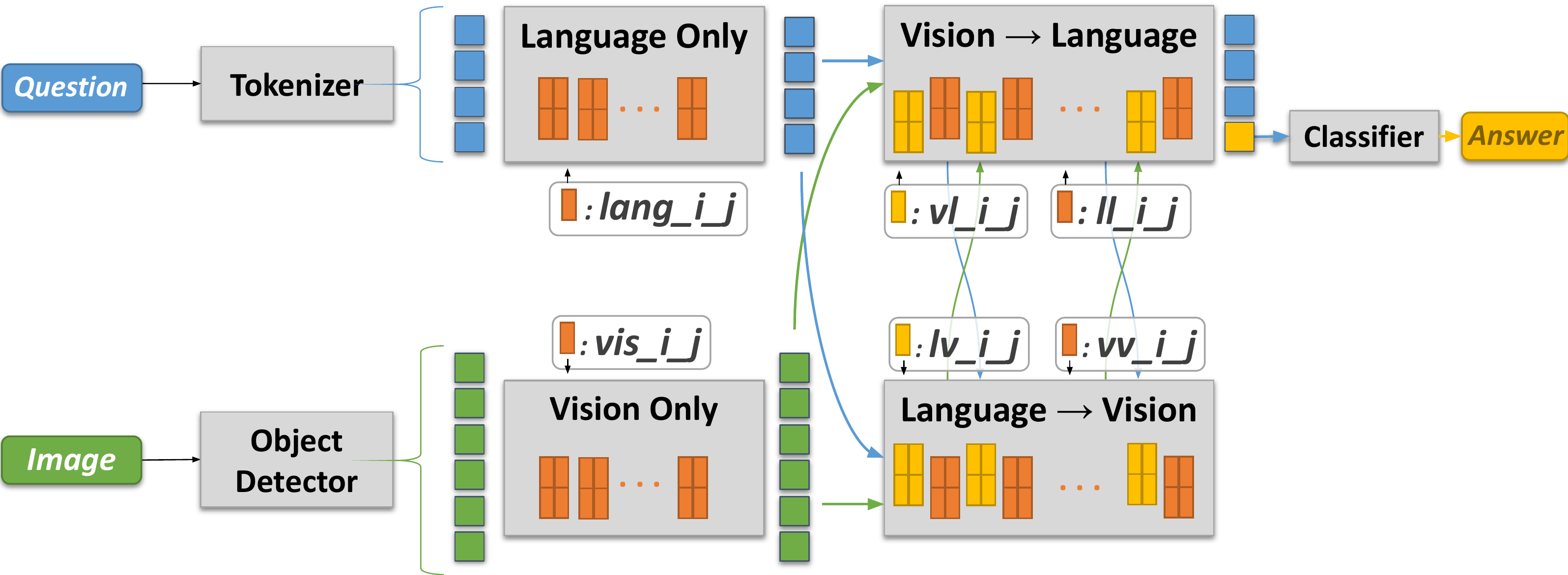}
    \caption{An Illustration of the VL-Transformer architecture used in the paper. Question and image are first tokenized and then encoded using vision (in green) and language (in blue) only transformers~\cite{vaswani2017attention}, followed by (bi-directional)  inter-modality transformers~\cite{Tan2019LXMERTLC}. The answer is predicted from the \emph{``CLS''} token. Yellow and orange rectangles represent, respectively, inter- and intra-modality attention heads. $i$ and $j$ are the layers and head indices used for naming attention heads through the paper.}
    \label{fig:transformer}
            \vspace{-.3cm}
\end{figure}

\subsection{Vision-Language (VL)-Transformers}
\label{sec:vlbg}
\added{
Transformers have been extended to reasoning on multiple modalities, in particular vision and language, through different types of layers:
Language-only and vision-only layers, referred to as \emph{intra-}modal layers, and language-vision layers, referred to as the \emph{inter-}modal ones. 
Fig.~\ref{fig:transformer} depicts the transformer architecture designed for VQA which we call ``\textit{VL-Transformer}''.}
\added{Each layer is named as \emph{X\_i\_j}, where \emph{X} denotes the layer type} (\textit{e.g.} \added{vision-only intra-modal layer, vision-language inter-modal layer, etc.) and $i$ and $j$ index, respectively, layer and head.}

\myparagraph{Intra-modality}
\added{
Both modalities are first processed in two independent streams (cf. Fig.~\ref{fig:transformer}):
 heads \emph{lang\_i\_j} encode question words,
and \emph{vis\_i\_j} heads encode visual objects detected in the image by an off-the-shelf object detector, a mainstream approach in  VQA~\cite{anderson2018bottom, Tan2019LXMERTLC}.
Visual embeddings are the concatenation of $2048$-dimensional object appearance embeddings and $4$-dimensional bounding box coordinates.
}

\myparagraph{Inter-modality}
\added{
Subsequent layers combine information between both modalities, c.f. Fig.~\ref{fig:transformer}, in a bidirectional way: from question words to visual objects in \emph{lv\_i\_j}, and vice-versa in \emph{vl\_i\_j} (\emph{lv} means `language to vision' while the opposite \emph{vl} means `vision to language').
This requires a minor, but essential, modification of the attention mechanism.
Intuitively, and a bit simplified, in vision-to-language heads, the language (word) embeddings are transformed by taking each word and checking its similarity to the full set of visual input objects, and vice-versa for language-to-vision heads. Details are given in supplementary materials.}
\added{
As shown in Fig.~\ref{fig:transformer}, each \emph{lv} or \emph{vl} attention head is immediately followed by an intra-modal attention head called, respectively, \emph{vv} or \emph{ll}.
}

\myparagraph{Predicting the answer}
\added{
The answer is produced by decoding the final representation of the ``\emph{CLS}'' token using a 2-layered neural network. It predicts a probability vector over the most frequent answers found in the training set, the answer with the highest score is then chosen.
}

\myparagraph{Training details}
\added{
For the experiments in this paper, we set the embedding size to $d{=}128$ and the number of heads per-layers to $h{=}4$. Following \cite{Tan2019LXMERTLC}, our model is composed of 9 language only and 5 vision only intra-modality transformers layers, and 5 language $\rightarrow$ vision and vision $\rightarrow$ language layers.
In addition to the VQA objective, we train the model parameters also on MS-COCO~\cite{lin2014microsoft} and Visual-Genome~\cite{krishna2017visual} images following the semi-supervised BERT~\cite{devlin2019bert}-like strategy introduced in~\cite{Tan2019LXMERTLC}. In particular, the model is trained to perform simple tasks such as recognizing masked words and visual objects, or predicting if a given sentence matches the question.
After pre-training on these auxiliary tasks, the model is fine-tuned on the GQA~\cite{hudson2019gqa} dataset with the VQA objective. Our VL-Transformer is a variant of the LXMERT model~\cite{Tan2019LXMERTLC}}, \cond{in line with the many works adapting BERT~\cite{devlin2019bert}-like pre-training to vision and language tasks~\cite{su2019vl, chen2020uniter, li2020oscar, NEURIPS2020_49562478, lu2019vilbert}.
}

\myparagraph{Discussion}
\added{
In this paper, we focus on the interpretation of the attention maps, as they contain crucial cues on the internal reasoning in transformers.
These maps highlight to what extent a given token has been contextualized by which neighbors, 
high attention $\bm{\alpha}_{ij}$ indicating  strong interaction between tokens $i$ and $j$.
We argue, that attention maps provide strong insights on how our the model handles interactions between the question the image.}

    

\section{Related Work}
\label{sec:related}

\added{Our work is related to building visual analytics tools for interpretability of Deep Learning. Our design targets in particular the study of attention maps from transformers models to grasp insights on their potential exploitation of bias. This section reviews previous work on the visual analysis of deep learning models, and a review of previous work from machine learning communities to tackle bias in VQA systems. }

\subsection{Visual Analytics for Interpretability}

In recent years, the visual analytics community has joined forces with machine learning experts and provided contributions improving the interpretability of deep neural networks~\cite{Hohman2019VisualFrontiers}, by providing insights on their inner workings. Those models are often considered black-boxes due to the large amount of parameters and data they manipulate to reach a decision. 
Prior work focused on the analysis of image processing models, known as convolutional neuronal networks, by exposing their gradients over the input images~\cite{zeiler2014visualizing}. This approach, enhanced with visual analytics~\cite{liu2016towards}, and provided glimpses on how the neurons of those models are sensible to different patterns in the input. More recently, CNNs have been analyzed through the prism of attribution maps in works such as Activation-atlas~\cite{carter2019activation} and attribution graphs~\cite{hohman2020summit}. 

On the other side, natural language processing (NLP) with recurrent neural networks, have also been explored through static visualization~\cite{Karpathy2015VisualizingNetworks} which provided insights, among others, on how those models can learn to encode patterns in sentences beyond their architectures in capacities. Interactive visual analytics works such as LSMTViz~\cite{Strobelt2017Lstmvis:Networks}, and RetainVis~\cite{kwon2018retainvis} have also addressed the interpretability of those models through visual encoding of their inner parameters, which can then be filtered and completed with additional information. Those parameters are collected during forward pass on models, as opposed to RNNbow~\cite{CashmanRNNbow:Networks}, which has the particularity to focus on visualizing gradients of those models through back-propagation during training.

\subsection{Interpretability of Attention}
More recently, models with attention~\cite{vaswani2017attention} increasingly gained popularity due to their improvement of state-of-the-art performance, and their attention mechanisms which may be more interpretable than CNNs and RNNs. The interpretability of attention models similar to the transformer models used in this work, initially designed for NLP, has also been addressed by visual analytics contributions. Commonly, in works such as~\cite{strobelt2018s, olah2016attention,vig2019multiscale}, the attention of those models is presented, in instance-based~\cite{Hohman2019VisualFrontiers} interfaces as graphs with bipartite connections that can be inspected to grasp how input words are associated with each other. Attention Flows~\cite{derose2020attention} addresses the influence of BERT pre-training on model predictions by comparing two transformers models applied to NLP. Similar to our work, such a tool displays an overview of each attention head with a color encoding their activity. Those methods are specific to NLP tasks. In this work, we address the challenges provided by the bi-modality of vision and language reasoning, and expand the interpretability of VQA systems which can rely on visual cues or dataset biases. \cond{Current practices of VQA visualization include attention heatmaps of selected VL heads based on their activation~\cite{li2020what} to highlight word/key-object associations, global overview heatmaps of attention heatmaps towards a specific token~\cite{cao2020behind}, and guided backpropagation~\cite{goyal2016towards} to highlight the most relevant words in questions. Following those works, \tool\ provides a visualization of every head's attention heatmaps and word/object associations, along with an overview of their activations.}



Our focus is on post-hoc interpretability~\cite{Lipton2016TheInterpretability}, \ie the analysis of a trained model's decision policy after-the-fact. Our approach relies on instance-based analysis, which displays inner model parameters with respect to current input. Such analysis is often combined with direct manipulation mechanisms designed to let users experiment with desired input conditions, like drawing the input~\cite{Carter2016ExperimentsNetworkb}. Our working hypothesis is that a transformer-based model's mode of operation, i.e. whether it is reasoning or exploiting dataset biases, is observable from its trained parameters, and in particular from attention maps, an intermediate representation dependent on parameters.

\subsection{Bias Reduction in VQA}
\label{sec:relatedworkbiasreduction}





Bias reduction has been addressed on the data side through cleaning and balancing. 
In particular, GQA~\cite{hudson2019gqa} focuses on semantics with the help of human-annotated scene graphs and automatically generated questions.
On the contrary, VQAv2~\cite{goyal2017making} dataset is crowdsourced, which leads to more natural questions, but includes cognitive and/or social biases~\cite{eickhoff2018cognitive} as well as annotation mistakes.
In addition, evaluation benchmarks have been specifically designed to identify the presence of bias dependencies. VQA-CP~\cite{agrawal2018don} proposes to evaluate models bias dependency by introducing distribution shifts between training and testing sets.
However, this approach has limitations to evaluate the decisions and biases exploitation of models as it emphasizes the diversity in answers rather than the reasoning itself. As result, random predictions can also contribute to improving a model's score on this benchmark~\cite{teney2020value, shrestha2020negative}.
On the other side, GQA-OOD~\cite{kervadec2020roses} undertakes a similar strategy while keeping the training set distribution untouched. Instead, the authors propose to evaluate the VQA performances on question-answer pairs regarding their frequency in the dataset. This makes it possible to evaluate both in- and out-of-distribution performances and, at the same time, estimating the reasoning ability of the system. Indeed, if a model's prediction is correct while the question-answer pair is rare, there are chances that the model has not used statistical biases. These datasets and evaluation benchmarks have lead to the conception of diverse bias-reduction methods. While an exhaustive survey of these methods is out of the scope of this paper, one can mention families of approaches such as training regularization~\cite{shrestha2020negative}, or counterfactual learning~\cite{abbasnejad2020counterfactual, gokhale2020mutant}. 

 \added{In this work, we define bias as the way how a model learns regularities or shortcuts from its training dataset, which may push it to provide answers which it frequently encountered as correct during training, without even considering information from the input image. Technically, we evaluate bias as introduced in~\cite{kervadec2020roses}. Hence, every question from the dataset is grouped using their topic (\eg ``furniture'') and function (\ie task extracted from the semantic of the question). Both the semantic and topic metadata are provided by the GQA dataset. Then for each kind of question, the frequency of their answers is computed. We estimate that the model exploits bias when it incorrectly predicts an answer which is among the 20\% most frequent answers for the given question, while its ground-truth answer is among the 20\% least frequent.}
While our proposed tool allows to partially open the black-box of transformer-based neural vision and language models in a very general sense, making it possible to inspect how they handle relationships between data items, we nevertheless specifically focus on the important problem of bias reduction.
\begin{figure}[t]
    \centering
    \includegraphics[width=\linewidth]{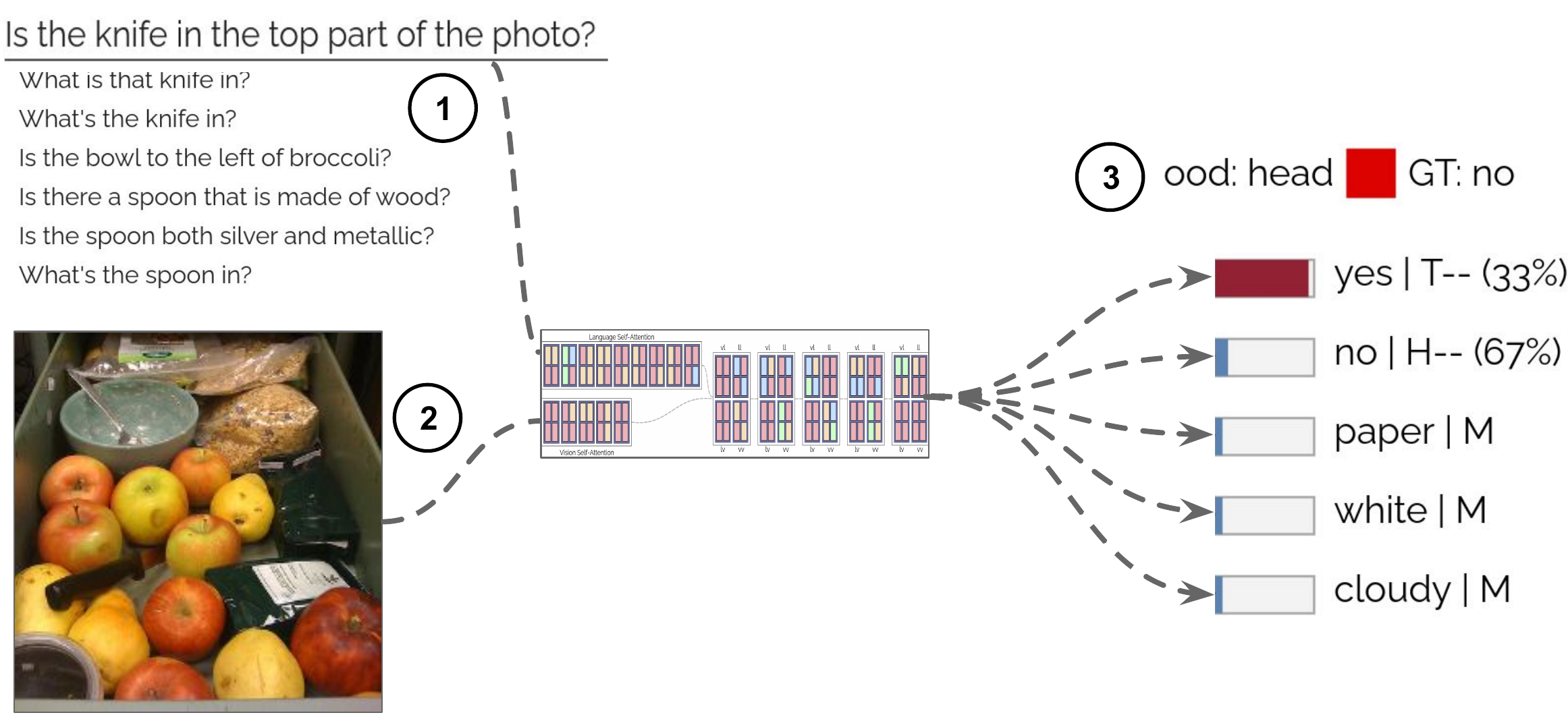}
    \caption{\label{fig:motivationfirst} \added{When asked \emph{``Is the knife in the top part of the photo''}~\ding{172} the tiny-LXMERT model, with the image of a knife at the bottom~\ding{173}, incorrectly outputs \emph{``yes''}~\ding{174} with more than 95\% confidence. While an exploitation of bias can be considered, we can observe that the answer \emph{``yes''} represents only 33\% of answers of similar questions over the complete dataset. Thus in-depth analysis of the attention of the model may be required to grasp what led to such a mistake.}}
            \vspace{-.3cm}
\end{figure}

\section{Motivating Case Study}
\label{sec:use-case}
This work was primarily motivated by growing concerns in the field over bias exploitation of models trained in large-scale settings, in particular when trained on very broad problems like vision and language reasoning~\cite{agrawal2016analyzing,goyal2017making, manjunatha2019explicit}. Our own contributions in this area include new benchmarks~\cite{kervadec2020roses} and additional supervision and regularization for neural models~\cite{kervadecweak}. 
Here we extend our previous efforts by providing a tool for instance-level visualizations and performing visual analytics on a single sample. This choice was ultimately taken when comparisons of different trained models through statistics failed to provide concrete answers on the sources of confusion and errors. Statistical models enabled us to drill down examinations to a certain minimal level of aggregation, for instance, linguistic function groups, but this kind of analysis was not fine-grained enough.




We illustrate the advantages and the power of instance-level visualizations with our contribution, \tool\, on the following case study, which is also available in video form in the supplementary material. It is based on the exploration of a tiny version of the state-of-the-art neural model LXMERT~\cite{Tan2019LXMERTLC} as described in sec.~\ref{sec:vlbg}. We provide it with the following input instance, \ie the image given in Fig.~\ref{fig:motivationfirst}(\ding{173}), and associated question \emph{``Is the knife in the top part of this photo?''}~\ding{172}. The correct ground truth answer is of course \emph{``No''}, but the baseline tiny-LXMERT model incorrectly answers \emph{``Yes''}~\ding{174}. We see the frequency of the different possible answers provided in the interface, and observe that the wrong answer \emph{``Yes''} is not the most frequent one for this kind of question as \emph{``No''} is the correct answer 67\% of the time, which does not provide evidence for bias exploitation. The objective is to use \tool~to dive deeper into the inner workings of the model.

A first step is to analyze whether the model is provided with all necessary information. While the input image itself does contain all clear pictures of the answer, the neural transformer model reasons over a list of objects detected by a first object detection and recognition module (Faster R-CNN~\cite{faster_rcnn2015}) which may output errors. In the rest of this paper, we will refer to this tiny-LXMERT as the \emph{noisy model}. 

\noindent
\textbf{Is the knife detected by the vision module?}
\tool\ provides access to the bounding boxes of the objects detected by the input pipeline. Each bounding box can be displayed superimposed over the input image along with the corresponding object label predicted by the object recognition module. We can observe that the key object \emph{``knife''} lacks a suitable bounding box or class label, it has not been detected. Since this object is required to answer the question for this image, the model cannot predict a coherent answer. However, the question remains why the wrong answer is \emph{``yes''}, corresponding to the presence of a knife.

\begin{figure}[t]
    \centering
    \includegraphics[width=0.7\linewidth]{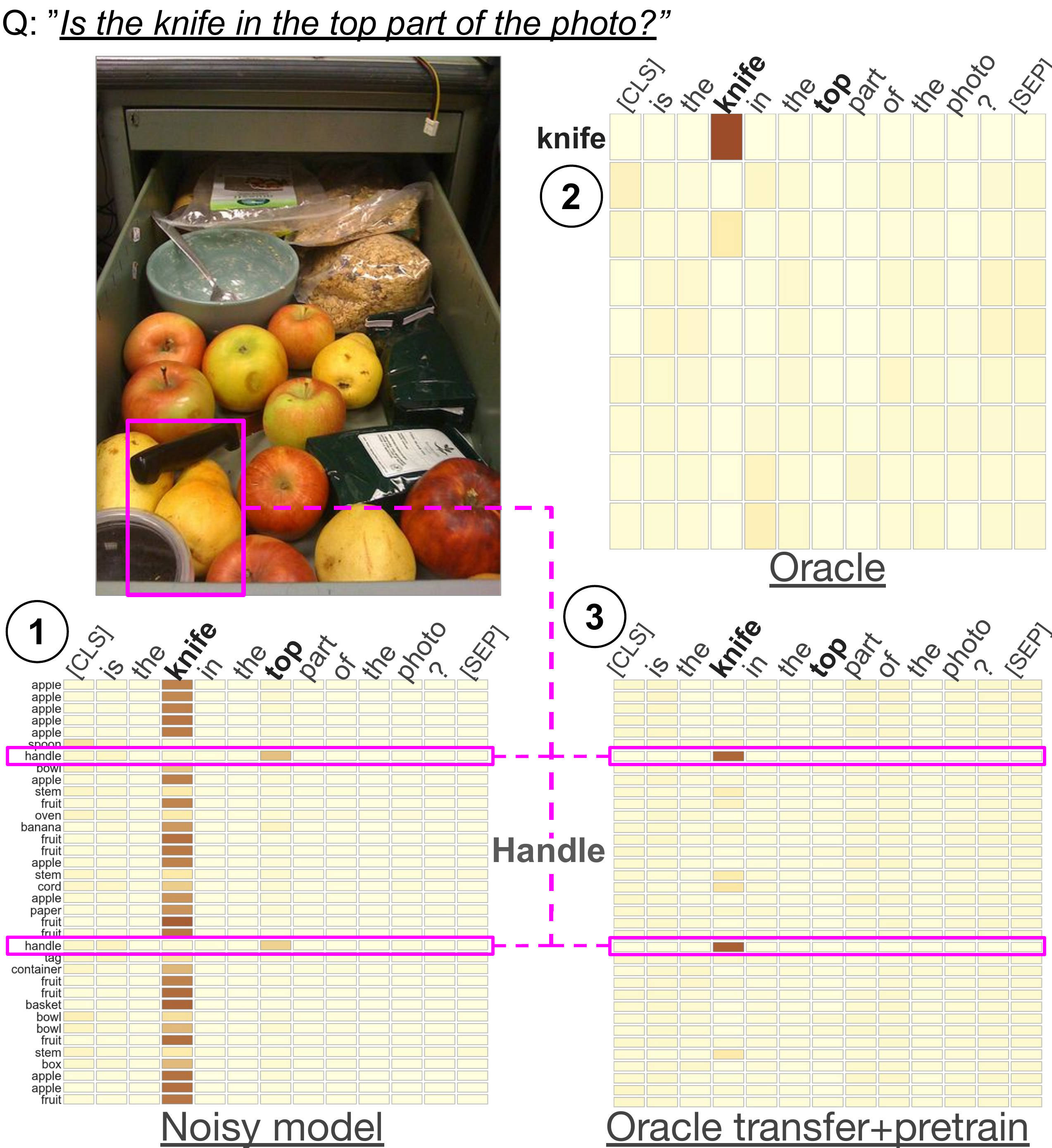}
    \caption{\label{fig:motivation}Visualization of a selected vision-to-language head and attention map for two different models. \ding{172} the \emph{noisy model} associates the \emph{``knife''} word with a large number of different objects, including fruit. \ding{173} the oracle model learns a perfect association between the word \emph{``knife''} and the \emph{``knife''} object; \ding{174} the oracle transfer model associates the word \emph{``knife''} with two different bounding boxes of type \emph{knife handle}, whose embeddings are sufficiently close for correct reasoning. Head selections are not comparable between models, we therefore checked for permutations.}
            \vspace{-.3cm}
\end{figure}

\noindent
\textbf{Can attention maps provide cues for reasoning modes?}
\tool\ focuses on attention maps, which are a key feature of transformer-based neural models, as they fully determine relationships between input items. Users can select different heads and explore the corresponding attention maps. For the example case, we are interested in checking the correspondence between the question word \emph{``knife''} and the set of bounding boxes, which should provide us with evidence whether the model was capable of associating the concept with the visual object in the scene, which is, of course, not sufficient for correctly answering, but a necessary step. This verification is non-trivial, however, since the model is free to perform this operation in any of the inter-modality layers and heads. \tool~allows to select the different heads, and we could observe that none of the heads provides a correct association. As an example, we can see the behavior of a head in Fig.~\ref{fig:motivation}~\ding{172}, which associates the word \emph{``knife''} to various objects, mostly fruits. No other head is found indicating a more promising relationship.




\noindent
\textbf{Is computer vision the bottleneck?}
From the example above, as well as similar observations in other instances, we conjecture that the computer vision input pipeline (notably, the imperfect object detector) is one of the main bottlenecks in preventing correct reasoning. To validate this hypothesis, we explored training an \emph{Oracle} model with perfect sight, which thus takes as input the ground truth objects provided by human annotation instead of the noisy object detections by a trained neural model. This improves the performance of the model considerably, reaching $\sim$80\% accuracy on the difficult questions with rare ground-truth answers, compared to $\sim$20\% for the standard model reasoning on noisy input. This particularly high difference in performance for questions with rare answers suggests a higher performance in correct reasoning of the oracle model. By loading this model into \tool, we observe in Fig.~\ref{fig:motivation}~\ding{173}, that there exists an attention map which associates the word \emph{``knife''} to a visual object \emph{``knife''}, which, we recall, is an object indicated through human annotation. This correct association is reassuring, but by itself does not yet guarantee correct reasoning --- further exploration is possible, but we will now concentrate on this problem of finding correspondences between words and visual objects and explore this question further.

\noindent
\myparagraph{Transferring reasoning modes}
Given these observations, we conjecture that a transformer model with perfect sight can learn modes of reasoning which are less biased than models trained on real but noisy input data, which is an insight we gathered from using \tool\ as a tool for visual analytics. Oracle models, on the downside, are not deployable to real-life problems, as they work on human annotations only, per definition. We explore a solution to transfer reasoning modes from oracle models to noisy input data, which can be done using knowledge transfer by parameter initialization~\cite{yosinski2014transferable}. In more detail, we pretain the Oracle model on ground-truth data, once it reaches convergence, use its weights as initialization for the model using noisy inputs. 

We load this model, which we call \emph{Oracle Transfer}, into \tool, with the objective of exploring whether reasoning modes have been transferred into a deployable model capable of providing answers given real input. Fig.~\ref{fig:motivation} shows the vision-to-language attention maps from the same layer as the ones explored above (\emph{Note: Each layer contains several attention heads, and these heads are not ordered. Reasoning associated with a given head in one model can correspond to the reasoning mode of a different head in another model. We cope with these possible permutations in our experiments by searching over all possible heads of a layer}). We can observe that the model's attention is drawn towards \emph{``handle''} visual objects, which are parts of the only knife in the picture. While the object classes \emph{``knife''} and \emph{``handle''} are logically different, we can conjecture that their vectorial feature embeddings are different but sufficiently similar to allow reasoning. More importantly, we can deduce that the model adapted to the absence of the knife object by relying on what is available, i.e. knife handles. This leads the model to correctly answering \emph{``No''}, as did the Oracle model on ground truth data, and in contrast to the baseline \emph{noisy model} without transfer --- see also the illustration in Fig.~\ref{fig:teaser}. We further describe this Oracle knowledge transfer, and improvements it provides to transformer-based VQA in a different associated publication~\cite{uscvpr2021}. This model will be used in evaluations performed by deep learning experts in section~\ref{sec:eval}. All models, noisy, oracle, and oracle transfer, can be tested and explored online in a prototype.

\section{\added{Design Goals}}

\cond{
Prior to the design of \tool, when working on an associated publication~\cite{uscvpr2021}, we conducted discussions with experts (co-authors of this paper), about their workflow to analyze bias in VQA systems. From those discussions, and literature review introduced in sec.~\ref{sec:related}, we distill their needs in the following four main themes of design goals for instance-based analysis.}



\begin{enumerate}[noitemsep,label=\textbf{G\arabic*},align=left,leftmargin=0cm,itemindent=.75cm]
    \item \added{\textbf{Examine the performances of each instance for a given model.} To investigate bias in VQA systems as introduced in sec.~\ref{sec:relatedworkbiasreduction}, it is first important to examine the model predictions, along with its confidence score with respect to its inputs. In order to be useful, those predictions need to be combined with the ground-truth, to estimate if the model is wrong, and how frequent is the ground-truth answer and predictions. Inputs need to be inspected as well as they may convey ambiguities that may be at the beginning of an explanation for a mistake. Finally, due to a large amount of data available for inspection, experts may prioritize inputs the more likely to be biased, \ie those with infrequent ground-truth answers and frequent predictions. }
    
    


    
    \item \added{\textbf{Browse the attention of the model for an instance}. 
    Analyzing the attention maps of LXMERT models is crucial for understanding what factors influenced its decision, and eventually whether or not the model attends to both language and vision. While visualizing individually each attention map is feasible, we aim at improving such an exploration, by contextualizing each attention head with their neighborhood (\ie other attention heads directly connected to it), and position within the model. This is relevant as attention heads get closer to the output, they both encapsulate previous attention, and may be more influential on models' decisions. In addition, experts may need to prioritize heads conveying salient attention, thus those heads need to be summarized and/or emphasized. Finally, for in-depth analysis, experts need to visualize the complete attention map and link each of their elements to human-understandable information, \ie words of the question and bounding boxes within the input image.}
    
    
    

\item \added{ \textbf{Link attention to language tasks.} Once a relevant attention head is observed by an expert, the user should be able to contextualize it with the rest of the dataset. In particular, experts are interested in evaluating whether or not this head is responsive to certain tasks provided by the semantics of questions (\eg find a color), or rather if the head is responsive to certain topics  (\eg clothing). Such information can be provided in the VQA training dataset, considered as categories.}


    
    \item  \added{\textbf{Explore alternative scenarios.} Once cues on how the model uses its attention to output a decision are gathered, the next step, is to test this knowledge by querying the model on altered input or parameters. Ultimately the experts desire to answers questions such as: ``would the model have a similar attention if the question were on another object of the image?'', or ``is this head or group of heads relevant for the final decision?''. This can be regrouped into two categories first, the possibility to ask free-from questions, and second the possibility to modify the model's attention. In order to be usable, and due to the number of queries an expert may need to execute, both those manipulations require to interact with the model in a reasonable amount of time, \eg less than a couple of seconds. }



\end{enumerate}

\section{Design of \tool}

We designed \tool, an visual analytics tool designed to facilitate in-depth analysis of the internal structure of transformers models applied to visual question answering. \tool\ implements attention heads and interactive heatmaps visualizations, and other interactions such as free-form question and pruning mechanisms, to assess whether a model resorts to reasoning or bias exploitation when answering questions.




\subsection{Workflow}
\label{sec:workflow}

\cond{Through iterative design resulting from frequent meetings with VQA experts co-authors of this work, we extracted the following workflow of use of \tool, based on their experience analyzing VQA systems, and the mantra \emph{overview first, zoom and filter, then details on demand}~\cite{shneiderman2003eyes}}:
\begin{enumerate}[noitemsep,align=left]
    \item the user picks and loads into the model an instance, informed by the likeliness of the question to be answered with bias; 
    
    \item the result of the model is presented both as attention heads (internal structure of the model), as well as the top-5 predicted answers; 
    
    \item the user then may interact with the model internal structure (\eg heads intensity, attention maps, etc.) which triggers updates to the statistical views;
    
    \item bounding boxes are displayed on the input image to reflect based on attention heat-map selections, showing how the model associates words and visual objects.
    
\end{enumerate}

\noindent     
\tool\ can also be used beyond this typical workflow for in real-time query of the model by asking user-defined questions, or pruning attention heads, as further detailed in Sec.~\ref{sec:interactions}.


\subsection{Visualization of Instances}

At its core, \tool\ is organized following an end-to-end approach, from model input to output. This is done to contextualize attention heads with their neighborhood and position with the model (\textbf{G2}). We added input selection to guide users in their exploration, and a visual summary of the attention of the model to keep it visually compact. 

\noindent
\myparagraph{Image ranking-by-feature~(Fig.\ref{fig:teaser}~\ding{172})} In order to ease user exploration over the complete dataset, \tool\ displays images in the top bar from left to right based on the likelihood of their questions to be answered using statistical biases. To do so, we classify questions using ground truth answers, as proposed in~\cite{kervadec2020roses}: top 20\% of the most frequent answer will be classified as \emph{Head}, as opposed to \emph{Tail} which describes questions with the least frequent answers.
We attribute a score to each image based on their \emph{Head}-questions/\emph{Tail}-questions ratio. The more an image has \emph{Tail}-questions over \emph{Head}-questions, the higher its score is. The underlying hypothesis is that frequent answers will be chosen more likely when a model tends to exploit biases (e.g. \emph{``Yellow bananas''}). Also, frequent questions are harder to analyze since if any bias is exploited by the model, it will answer correctly. Answers and their frequencies are displayed in~(Fig.\ref{fig:teaser}~\ding{173}~right) following design goal (\textbf{G1}).

\noindent 
\myparagraph{Instance view~(Fig.\ref{fig:teaser}~\ding{173})} This view, inspired by VL-Transformer representations illustrated in Fig.~\ref{fig:transformer}, is the root of any analysis of attention maps using \tool\ (\textbf{G2}). It matches the internal data flow through the internal structure of the model, from left to right: the input image/question pair, layers and heads with intra-modality layers first, and finally the answer output distribution (encoded as horizontal bars). A particular design decision was to display all attention maps at once, using a single-colored rectangle encoding the attention intensity as k-number~\cite{ramsauer2020hopfield} (see next paragraph for details). In the \emph{Instance view}, attention heads can be selected with a mouse-over interaction as illustrated in~Fig.\ref{fig:teaser}~\ding{174} in order to provide details on demand, by displaying the corresponding attention heat-map~Fig.\ref{fig:teaser}~\ding{175} and head statistics.

\noindent 
\myparagraph{Visual summary} Our model uses $136$ attention maps with dimensions varying with respect to the number of words in the question and bounding boxes provided. As displaying all of those matrices would prevent experts to analyze them in a reasonable amount of time, we rely on summarizing each of them to a single scalar. Such a scalar, referred to as \emph{k-number}~\cite{ramsauer2020hopfield}, represents the normalized amount of tokens per row summed up to reach a threshold of $90\%$ of energy. A \emph{k-number} close to $0$ indicates that the corresponding row has peaky attention focusing on only one column (as seen in Fig.~\ref{fig:motivation}~\ding{172}), and a high \emph{k-number} encodes a uniform attention (as in Fig.~\ref{fig:motivation}~\ding{173}). Then we combine each of those \emph{k-number} together using either \emph{min}, \emph{median}, or \emph{max} functions. Such functions can be selected in \tool\ by users, depending on the attention maps intensity they want to investigate. \tool\ provides this interaction because for a head to have a low \emph{k-number}, the majority of its rows needs to be highly activated. This can shadow attention maps with less than half of their rows with peaky attention. In \tool, the \emph{k-number} is discretized and color encoded in $4$ categories as it follows: \raisebox{-.25\height}{\includegraphics[height=.40cm]{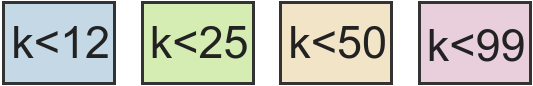}}. \added{The decision to use a logarithmic discretization and color encoding, as introduced in~\cite{ramsauer2020hopfield}, was done to emphasize peaky attention maps ($k < 12$) that need to be prioritized for analysis. In addition, this discretization, used throughout \tool, is particularly useful to regroup instances in head-stat view~Fig.~\ref{fig:teaser}~\ding{175} and select them by clicking on the corresponding square (legend on the right of~ \ding{173} in~Fig.~\ref{fig:teaser}) for \emph{Head pruning} further detailed in section~\ref{sec:interactions}. Finally, this reduces the learning curve of \tool, as experts are familiar with this discretization of \emph{k-numbers}.}

\subsection{Visualization of Selected Heads}

\tool\ provides details for a selected head in the Instance view using the attention map and head statistics.

\noindent 
\myparagraph{Attention Maps (Fig.\ref{fig:teaser}~\ding{176})} are represented as heat-maps, with cell colors encoding the attention intensity over a sequential, single hue scale \added{from \emph{no attention} ( \ie $0$) in beige, to \emph{full attention} (\ie $1$) in brown}. This matrix-based approach contrasts with bipartite connections representations found in Seq2Seq-Vis~\cite{strobelt2018s} and BertViz~\cite{vig2019multiscale}. We use matrices as they provide a clutter-free representation of attention and they can display multiple heads at once by aggregating connections. However, it may suffer from visual clutter issues as the number of words grows. Seq2Seq-Vis tackles this issue by using interactions to hide graph lowest connections (\ie lowest attention). In our case, we argue that visualizing the attention of one head at a time can lead to a better understanding of its function in the model. Such an exploration of head functions is further detailed in Sec.\ref{sec:eval}.





\noindent 
\myparagraph{Head Statistics (Fig.\ref{fig:teaser}~\ding{175})} are represented using three charts. A vertical area chart (leftmost chart) represents the distribution of \emph{k-numbers} of the selected head over the complete validation dataset (around $1500$ image/question pairs). The vertical axis encodes the values of \emph{k-numbers}, while the horizontal axis encodes the density of the corresponding \emph{k-number}. The current \emph{k-number}, for the selected head, which corresponds to the image/question pair loaded in \tool, is represented as a horizontal red bar positioned on the vertical axis. This area-chart provides insights such as the detection of useless heads with constant high \emph{k-number} which can be reduced to calculation on average overall items instead of selecting specific items. In contrast, heads with constant low \emph{k-number} can be interpreted as conveying key information. More specialized heads, with bi-modal \emph{k-number} distributions, can also be observed. Two stacked bar-charts represent the \emph{k-numbers} of the selected head grouped by question operations (\textbf{G3}). Question operations are ground truth information provided by the GQA dataset~\cite{hudson2019gqa} describing the semantic reasoning operation of asked questions (\eg select, query..). Each bar is associated with an operation, and its length encodes how many questions in the dataset are using the corresponding operation.






\begin{figure}[t!]
     \vspace{-.2cm}
    \centering
\includegraphics[width=0.6\linewidth]{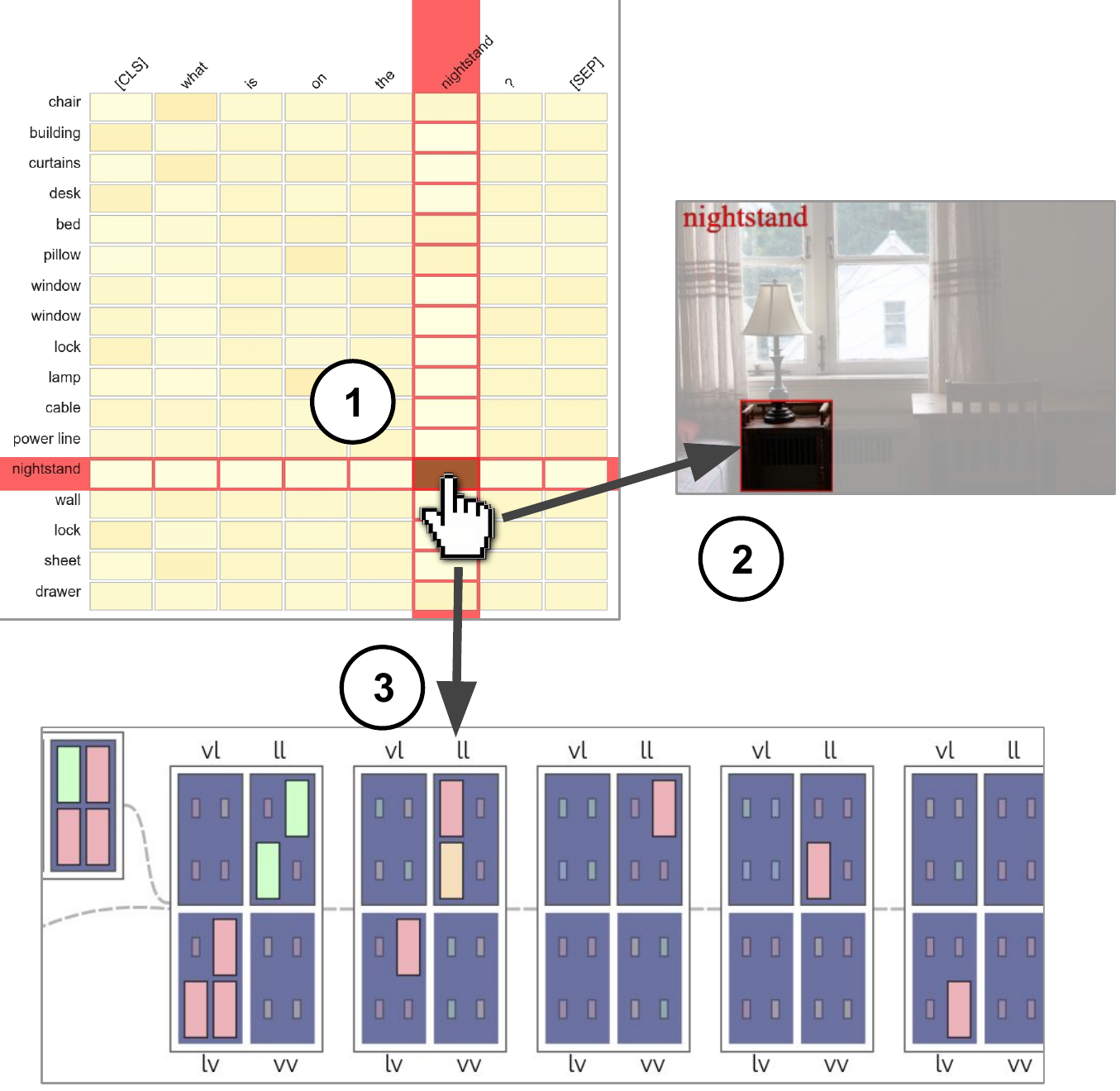}
    \caption{Hovering the mouse over a cell of the attention maps~\ding{172} filters the corresponding object bounding box in the input image~\ding{173}. While clicking on this cell filters attention heads in \emph{instance-view} to display those within which the selected cell is highly activated.}
    \label{fig:HeadFilter}
            \vspace{-.4cm}
\end{figure}


\subsection{Interacting with Models}
\label{sec:interactions}

\tool\ also includes features, complementary to the usual workflow introduced in Sec.\ref{sec:interactions}, designed to investigate hypotheses on reasoning. 

\noindent
\myparagraph{Free-form questions}
By default, \tool\ loads the GQA dataset~\cite{hudson2019gqa} to provide images and questions. But at any time, users can type and ask free-form open-ended questions (\textbf{G4}). Such an interaction allows investigating the model's bias exploitation. For instance, when asked the following question from the GQA dataset \emph{``Is this a mirror or a sofa''}, the model correctly outputs ``\emph{mirror}''. However, when asked the following user-inputted question \emph{``Is there a mirror in this image?''}, the model fails and outputs \emph{``no''}. This suggests that the model might have exploited biases when it answered the first question, which is supported by the fact that in the GQA dataset, \emph{``mirror''} is the correct answer to the question \emph{``Is this a mirror or a sofa''} in 85\% of all cases.

\noindent
\myparagraph{Head filtering (Fig.\ref{fig:HeadFilter})}
As shown in Fig.\ref{fig:HeadFilter}~\ding{172}, attention heat-maps feature two interactions. First, hovering with the cursor a row, a column, a cell, which corresponds to an object in the image, automatically displays its corresponding bounding box along with its label over the input (Fig.\ref{fig:HeadFilter}~\ding{173}). The user can also, by clicking on a row, column, or cell, filter attention heads to only keep the ones in which the corresponding clicked element has attention above a threshold. For rows and columns which contain multiple attention values, such a filtering process will merge those values by using one of three functions \emph{min}, \emph{median}, and \emph{max} depending on a user selection. The result of such a filtering, as displayed in Fig.\ref{fig:HeadFilter}~\ding{174}, occurs in the \emph{instance view} in which the size and opacity heads that do not match with the user's query are reduced while the others are preserved. Such interactions facilitate seeking for heads in which a specific association is expected \eg a word in the question with an object of the image required to answer.

\noindent
\myparagraph{Head pruning}
Users can select attention heads by clicking on them in the \emph{instance view}, or by their \emph{k-number} category. Such a selection can then be used to prune the corresponding head for the next forward of the model. Pruning here means that the attention head does not perform any focused attention, but uniformly distributes attention over the full set of items (objects or words). Each row of a pruned attention map is thus the equivalent of an average calculation. At any time, users can request a new forward pass of the model by clicking on the top left button ``ask'' (\textbf{G4}), which allows to see the effect of the configured pruning on the model's predictions. This can be used in order to test hypotheses on attention head interpretations as explored in Sec.~\ref{sec:eval_LV}.


    

\section{Implementation}

The GQA~\cite{hudson2019gqa} standard dataset provides question/image pairs along with their answers, the ground truth of bounding boxes, and semantic descriptions of questions. By default, \tool\ provides around $1500$ question/image pairs, but as images are loaded progressively when users request it, such a quantity can be increased without affecting performances. The models have been trained on a significantly larger amount of training data (about $9$M image/sentence pairs), which is different from the validation data on which the performance is evaluated. The user interface of \tool\ is implemented using D3~\cite{Bostock2011D3:Documents}, and directly interacts with transformer models implemented in Pytorch~\cite{NEURIPS2019_9015}, using JSON files through a python Flask server. As the possibility to ask free-form questions offers an infinity of possible combinations, each question/image pair is forwarded through the model in a plug-in fashion, \ie without altering the model and its performances. 



\section{Evaluation with Domain Experts}
\label{sec:eval}

We conducted a user study with $6$ experts with experience in building deep neural networks, who were not involved in the project or the design process of \tool. We report on their feedback using \tool\ to evaluate the decision process of the \emph{Oracle transfer model}, \cond{with 57.8\% accuracy on GQA},  introduced in Sec.\ref{sec:use-case} on several provided problem instances, as well as insights they received from this experience. \cond{Hypotheses drawn from single instances cannot be confirmed or denied, but as illustrated in the following sections, such a fine-grained analysis aims to provide cues (often unexpected) that can later be explored through statistical evidence outside of \tool.}

\subsection{Evaluation Protocol}
For each expert, we conducted an interview session lasting on average two hours. Sessions were organized remotely and began with a training on \tool, showing step-by-step how to analyze attention maps. During this presentation, experts were able to ask questions. The study then began with questions on $6$ problem instances, \ie image/question pairs loaded into \tool\ in a browser window on participants' workstations. Those instances were balanced between the prediction failures and successes, head or tail distributions of question rarity as described in sec.~\ref{sec:relatedworkbiasreduction}, as well as our estimation on whether the model resorts to bias for this instance grasped using \tool. \added{\tool, configured as conditioned during evaluations is accessible online at:}~\url(https://theo-jaunet.github.io/visqEval/). The model outputs were hidden and the experts were asked to use \tool\ to provide an estimate for two different questions: (1) will the model predict a correct answer, (2) what will it be?, and (3) does it exploit biases for its prediction, or does it reason correctly? 
During this part of the interview, experts were asked to explain out loud what lead them to each decision. Once those questions were completed, post-study questions were asked on the usability of \tool, such as \emph{``Which part of \tool\ is the least useful?''}, and \emph{``What was the hardest part to understand?''}.

\noindent    
\myparagraph{Results}
\cond{The ability of users to predict failures and specific answers of VQA systems has already been addressed through evaluation~\cite{Chandrasekaran2018} under different conditions. The experiment closest to ours is question+image attention~\cite{lu2016hierarchical} with instant feedback --- similarly to ours, users were asked to estimate whether a model will predict a correct answer when provided with attention visualizations of the model, and reaching a similar score of $\sim$75\% accuracy. The difference is that in~\cite{lu2016hierarchical} attention is overlaid over the visual input, whereas our attention maps allow to inspect reasoning in a more detailed and fine-grained manner, and not necessarily tied to the visual aspects. The similarity in results changes when users are asked to provide the specific answer predicted by the model: this accuracy drops to 61\% in our case, and to 51\% in~\cite{Chandrasekaran2018}. While our results are promising, they cannot be directly compared to their results due to the different pool and amount of users. Future work will address studies on a larger number of human experts.   
}

More importantly, our work focuses on qualitative results of bias estimation in which experts obtained a precision of 75\% on whether the model exploited any bias. We extracted the ground truth estimate by comparing the rarity of the question, following~\cite{kervadec2020roses}. Supplementary material provides more details on the evaluation protocol and experts' performances. These results are encouraging, as they provide a first indication that the reasoning behavior of VL models can be examined and estimated by human users with \tool. While 75\% of performance reasoning vs. bias is not a perfect score, it is also far away from the random performance of 50\%, which is important given the large capacity of these models, which contain millions of trainable parameters. 


\begin{figure}[t!]
     \vspace{-.25cm}
    \centering
\includegraphics[width=0.7\linewidth]{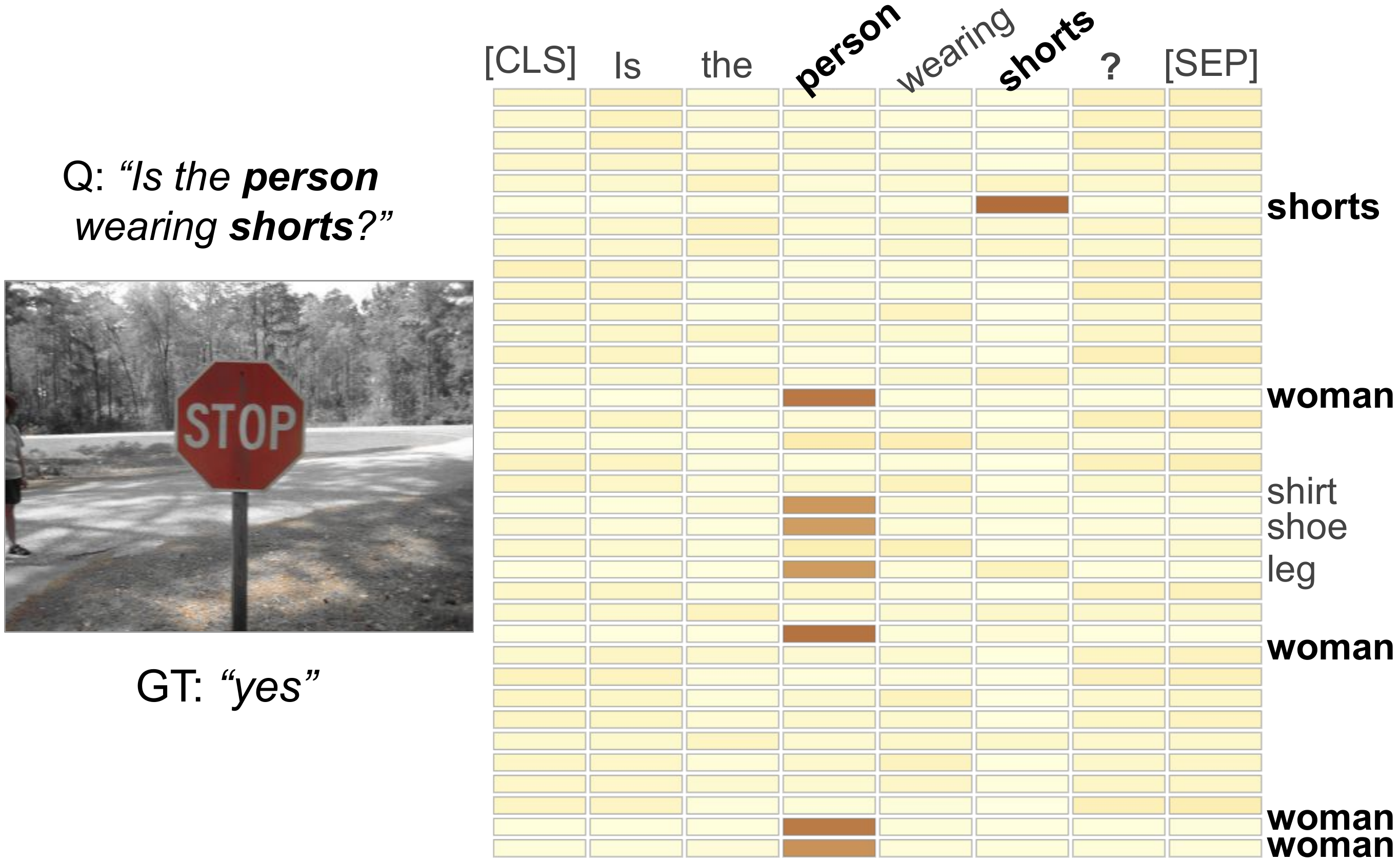}
    \caption{When asked \emph{``Is the person wearing shorts?''}, the \emph{oracle transfer} model successfully answers \emph{``yes''}. It can be observed in its first Language-to-Vision attention maps, that the word \emph{``shorts''} (column) is strongly associated with the object \emph{``shorts''} (row). The same phenomenon is also observed for the word \emph{``person'}, strongly associated with objects labeled as \emph{``woman''} among others.}
    \label{fig:caseLV}
    \vspace{-.4cm}
\end{figure}

\subsection{Object Detection and Attention}
\label{sec:eval_LV}
To provide an answer, a model must first grasp which objects from the image are requested and thus are essential to focus on. Such an association needs to occur early in the model as those objects are needed for further reasoning. The experts widely observed high intensity in the first language-to-vision (LV) layer. As illustrated in Fig.~\ref{fig:caseLV}, when asked \emph{``Is the person wearing shorts?''}, the attention map~\emph{LV\_0\_1} has peaky activations in the columns \emph{``person''} and \emph{``shorts''}. This can be interpreted as the model correctly identifying with its self-attention for language that those two words are essential to answer the given question. In Fig.~\ref{fig:caseLV}, the word~\emph{``person''} is associated with the bounding boxes labeled as \emph{``woman'',``shirt'',``shoe'',``leg''}, while the word~\emph{``shorts''} is associated with the~\emph{``shorts''} bounding box. Based on this observation, all experts concluded that the model correctly sees the required objects, and more broadly over the evaluation instances, that the first LV layer might be responsible for the recognition of objects with respect to the question. One of the experts mentioned that therefore, \emph{``if we don't see a good word/bounding-box association here, the model can hardly cope with such a mistake and might exploit dataset biases''}. In order to verify such a statement, we pruned the four heads in this LV layer, to observe how the model would behave with no association in them. From such pruning, we observe that the following vision-to-language (VL) layers have lower attention distributions, close uniform in some cases. In addition, after pruning, the model's prediction wrongly switched from \emph{``Yes''}, a rare answer (in Tail), to \emph{``No''}, the most frequent one. 






\subsection{Questions with Logical Operators}

\begin{figure}[t!]
     \vspace{-.25cm}
    \centering
    \includegraphics[width=0.8\linewidth]{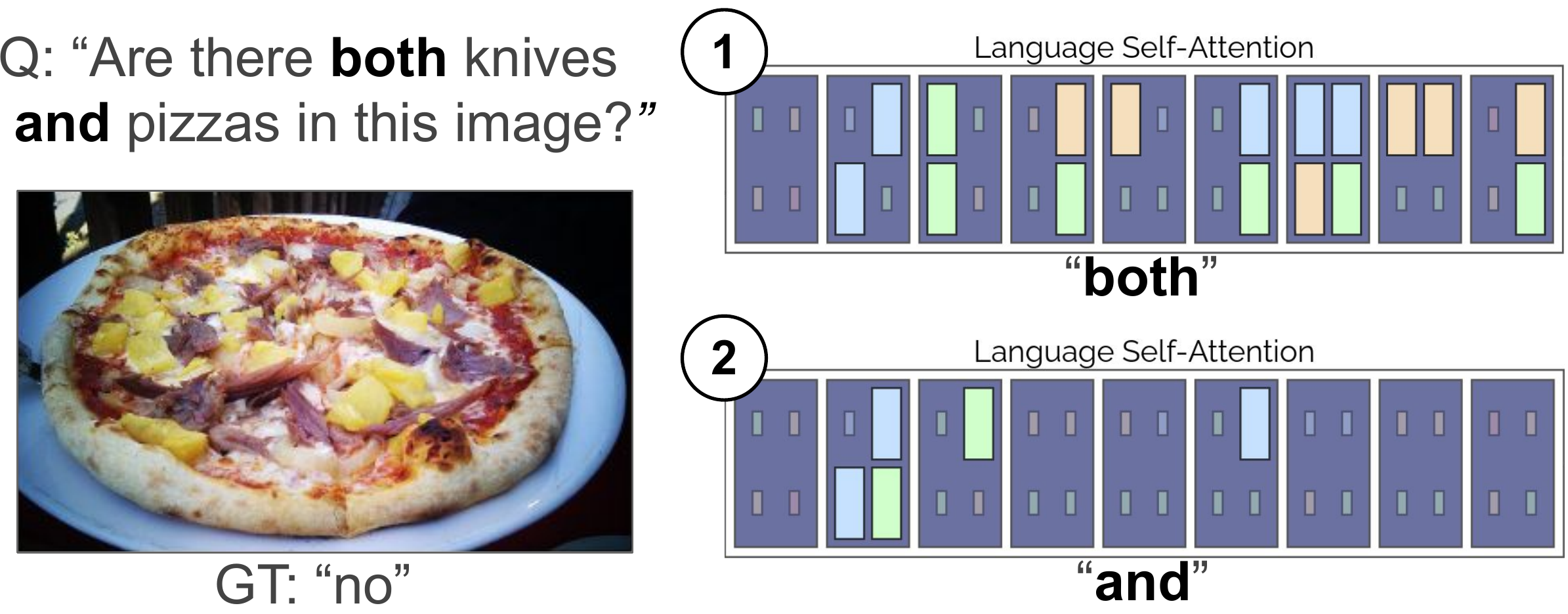}
    \caption{When asked \emph{``Are there both knives and pizzas in this image?''}, the \emph{oracle transfer} model fails and answers \emph{``yes''}. By filtering heads associated with a selected word, we can observ that language self-attention heads are more responsive to the word \emph{``both''}~\ding{172}, as opposed to the word \emph{``and''}~\ding{173}. }
    \label{fig:logical}
        \vspace{-.4cm}
\end{figure}

During the evaluation, experts were shown two instances with questions containing the word \emph{``and''}. Such instances are interesting because, as one of the experts mentioned, \emph{``this word has a lot of importance is this question''}. To answer correctly, the model needs to grasp that it must analyze the image over two different aspects. With the image, illustrated in Fig.~\ref{fig:logical}, and asked \emph{``Are there both knives and pizzas in this image?''}, the model fails and answer \emph{``yes''}, the most frequent answer despite having no knife in the picture nor provided bounding-boxes. However, when asked \emph{``Are there knives in this image?''} the model correctly answers~\emph{``no''}. This suggests that the model failed to grasp the meaning of the keyword \emph{``and''}, and thus that the self-attention language heads might associate wrong words. Also, swapping the terms \emph{``knives''} and \emph{``pizzas''} in the question, yields the correct answer, \ie \emph{``no''}. This indicates that the model ignores the first term when questions contain the operator \emph{``and''}. Using the head-filtering interaction, we can observe that in self-attention heads, the word \emph{``and''} has little to no attention. Instead, the word \emph{``both''} has peaky attention scattered across most of self-language layers, and some language-to-language heads. Pruning those $19$ heads makes the model correctly yield \emph{``no''}, regardless of the order the words \emph{``knives''} and \emph{``pizzas''} are in the question. Such a behavior can be observed over our evaluation dataset, in which $34$ questions have the keyword \emph{``and''}. On those questions the model, without pruning, 
can provide a correct answer 62\% of cases, up to 64\% with the two words around \emph{``and''} are swapped. In opposition, while having the $19$ attention-heads with peaky attention for the word \emph{``both''} pruned, the model reached an accuracy of 76\%, down to 74\% with words around \emph{``and''} swapped. Thus, in the worst case, this pruning of the $19$ attention heads illustrated in Fig.~\ref{fig:logical} is responsible for an improvement of 10\% on question contain the operator \emph{``and''}.

\subsection{Vision to Vision Contextualization}

\begin{figure}[t!]
    \centering
    \includegraphics[width=0.9\linewidth]{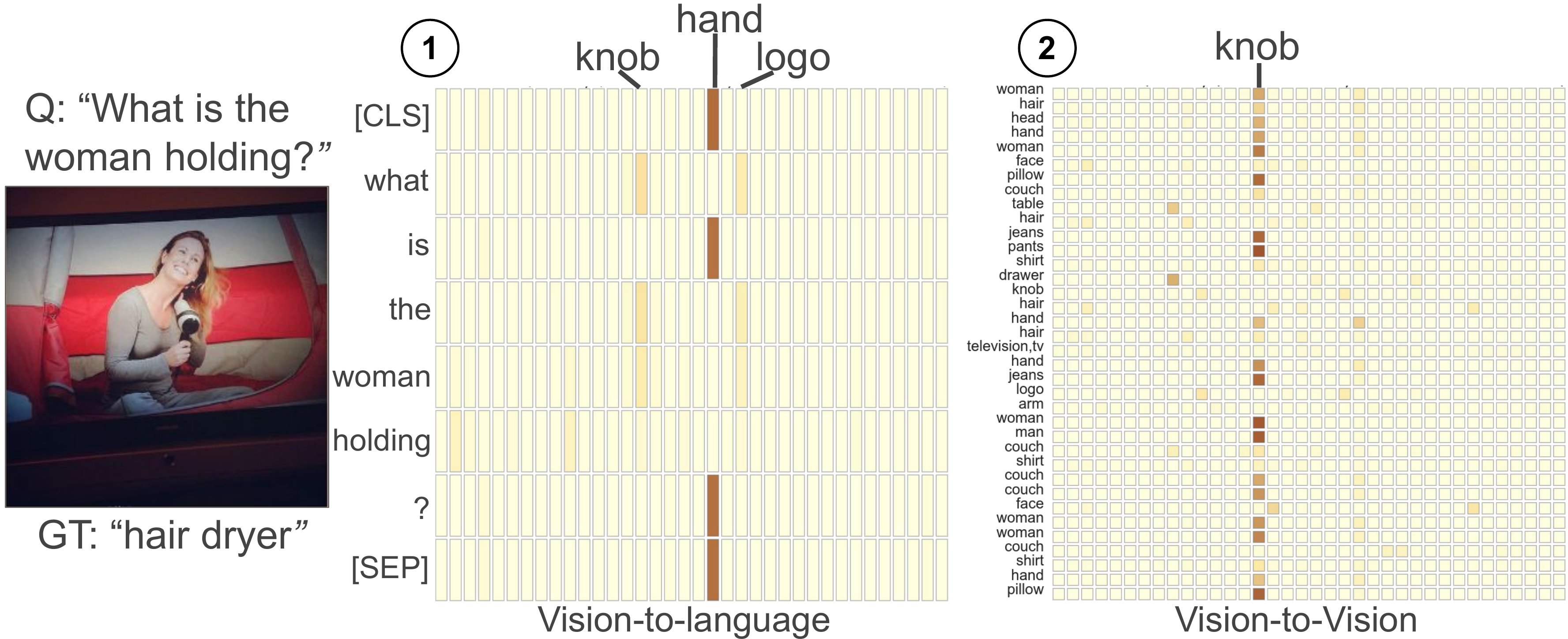}
    \caption{Without any\emph{"hair dryer"} provided by the object detector, the \emph{oracle transfer} associates in its vision-to-language~\ding{172} the object \emph{``hand''} with the words \emph{\{``[CLS]'',``is'',``?'',``[SEP]''\}}. While vision-to-vision focuses on a \emph{``knob''} object~\ding{173}.}
    \label{fig:word}
        \vspace{-.3cm}
\end{figure}

When asked \emph{``What is the woman holding?''}, with the image in Fig.~\ref{fig:word}, the \emph{Oracle transfer} model fails and outputs \emph{ ``remote-control''}, a frequent answer, instead of \emph{ ``hair dryer''}. This can be interpreted as the model exploiting a statistical bias from its training dataset. However, in such a dataset, \emph{ ``remote-control''} is not among the $10$ most common answers to this question. This raises the question of what leads the model to output such an answer. During evaluation on this instance, experts noticed that the object detector failed to provide a \emph{ ``hair dryer''} object. Similar to the use case given in Sec.~\ref{sec:use-case}, such a mistake forces the \emph{Oracle transfer} to draw its attention towards other bounding boxes related to the missing object. In this case, as observed by experts, a majority of the vision-to-language reached their highest association between the word \emph{``holding''} and bounding boxes labeled as \emph{``hands''}. Such an association is expected as held objects are directly related to hands, and no\emph{ ``hair dryer''} bounding box is provided. Among those bounding boxes, we can observe the presence of one labeled as \emph{``television''}, and another as \emph{``knob''} which are associated to \emph{``holding''} and \emph{``woman''} in both vision-to-vision\_2\_2 and early vision-to-language layers. This suggests that those heads might have influenced the model's predictions towards \emph{ ``remote-control''} instead of the most common dataset bias. This can be confirmed by pruning those heads which yields a more frequent answer: \emph{``cell phone''}. In addition, one of the experts also highlighted that those attention heads had a high association with the tokens \emph{``[CLS]''}, \emph{``is''}, \emph{``?''}, and \emph{``[SEP]''}. Which the expert interpreted as \emph{``the model correctly transferred the context of the question''}.

\section{Discussions, Limitations and Future Work}

\myparagraph{Usability of \tool} Overall, \tool\ was positively received by experts, during discussions at the end of interviews, they expressed that \tool~is \emph{``well designed''}, \emph{``complete''}, and \emph{``particularly useful as VQA transformers are hard to interpret''}. Two out of the six experts confessed that they felt \emph{``overwhelmed at first''}, but gradually \emph{``grasped where to look''}, and getting \emph{``used to interactions''}. Four out of six experts stated that the \emph{head-statistics}~\ref{fig:teaser}~\ding{175} is the least useful feature implemented in \tool, as it felt \emph{``hard to understand''}. One expert mentioned that such a view is \emph{``useful in theory, but less usable in practice''}. However, the rest of the experts mentioned that the \emph{head-statistics} view helped them while analyzing attention to \emph{``see if heads acted out of their distribution''}. Experts were unanimous, the attention heat-map, and in particular, its interactions, is the most useful feature of \tool, as it \emph{``provides a lot of information'}, and \emph{head filtering} \emph{``speeds up the analysis''}. One of the expert used this interaction on the first language to vision layer as an entry-point to grasp if the model had every information required to correctly answer the given question. 

\myparagraph{Expert Suggestions} As expressed by experts during evaluation, the main limitation of \tool\ is how instances can be selected by users. Currently, such selections are handled by a bar at the top of the tool~(Fig.\ref{fig:teaser}~\ding{172}), displaying images ordered from left to right based on the likelihood of their questions to be answered using statistical biases. However, during interviews, experts mentioned their desire to quickly switch between similar instances in order to grasp if a behavior can be seen across different cases. To do so, experts suggested that such similarity could be measured at three levels: switch to an instance with the exact same question, switch to a similar image, and switch to an instance in which a selected head has similar attention distribution. In addition, currently in \tool, it is difficult to evaluate how influential a head is over the model's output. To tackle such an issue, experts proposed to encode this information in the representation size of the \emph{instance-view} attention heads. 
The impact of each head over the model's input could be retrieved through back-propagation, in particular by calculating the gradient of the model outputs with respect to a statistic of the attention head, for instance, its k-number.
In addition to addressing those limits and experts' suggestions, we plan as future works to adapt the usage of \tool\ on other problems involving transformers, such as machine translation, and to further investigate the role of each attention heads as done in~\cite{voita2019analyzing}.

\added{
\myparagraph{Scalability} To date, the largest model loaded in \tool ~is LXMERT with $12$ attention heads per block and $768$ hidden dimensions. Using \tool, we noted that this model has a similar behavior as the tiny-LXMERT~\cond{(see supplementary material), and can outperform LXMERT with auxiliary losses~\cite{kervadec2021supervising}}. In addition, LXMERT is less accurate than the tiny-oracle model we used for evaluation, on questions with rare answers--\cond{\ie those that may be the most susceptible to convey biases}. While inspecting this model with \tool, we noted that the more the number of heads increases, the more summary and filtering of heads became relevant for faster analysis. However, our visual encoding of heads may become tedious to use as the number of heads increases, and the amount of pixels allocated per head decreases. A workaround can be to increase the designated space of the model in instance-view, but ultimately, \tool\ will be limited by screen real-estate. \cond{Thus, larger models (\eg UNITER-large~\cite{chen2020uniter}), may need their heads to be filtered (\eg by k-numbers) to avoid being overwhelming, before being displayed in \tool.}
Similarly, heatmaps of attention may suffer from the same limitation. However, we decided to use them for attention as the maximum visual object (36 in most VQA systems using a Faster R-CNN), and the average of words per question is known beforehand. For larger sets of inputs, alternative designs such as bipartite graphs combined with aggregation methods to hide elements under a threshold of attention may be more space-efficient. However, we argue that this is a trade-off as displaying every bounding boxes and their attention may be relevant to evaluate a prediction (as depicted in sec.~\ref{sec:use-case}).}


 \cond{
 \myparagraph{Generalization} This work focuses on detector-based VL transformers such as LXMERT, however other VQA systems may include single stream
 models such as~\cite{chen2020uniter,su2019vl}. While \tool\ is applicable to both single and 2-stream models, we decided to focus on the latest during evaluation. Those models are considered by experts as more interpretable because self-attention layers for language and vision can be observed separately, and there are no significant differences in performance between both architecture~\cite{bugliarello-etal-2021-multimodal}. VQA systems' state-of-the-art often alternate between detector-based approaches (\eg VinVL~\cite{zhang2021vinvl}) and pixel models~\cite{huang2020pixel}. However, at its current state, \tool\ is not applicable to the latter type, as it would only require adapting to the higher number of visual tokens (depending on the granularity of the method) which raises the challenge of scalability of heatmaps discussed above. Future works can address this through aggregation methods to reduce the number of tokens displayed (\eg by dividing input images in regions).
 Finally, our work mainly focuses on insights on reasoning skills grasped from models on the GQA dataset. This decision was taken because it has been argued~\cite{hudson2019gqa} that it involves a larger variety of reasoning skills (spatial, logical, relational, and comparative) than in a dataset with where questions were annotated by humans directly (\eg VQA2~\cite{goyal2017making}) which contains questions targeting difficult external knowledge (\eg a name of a baseball team). Despite not being presented in this work, \tool\ can directly be used on the VQA2 dataset.}
 

 \begin{figure}[t!]
      \vspace{-.25cm}
    \centering
\includegraphics[width=0.8\linewidth]{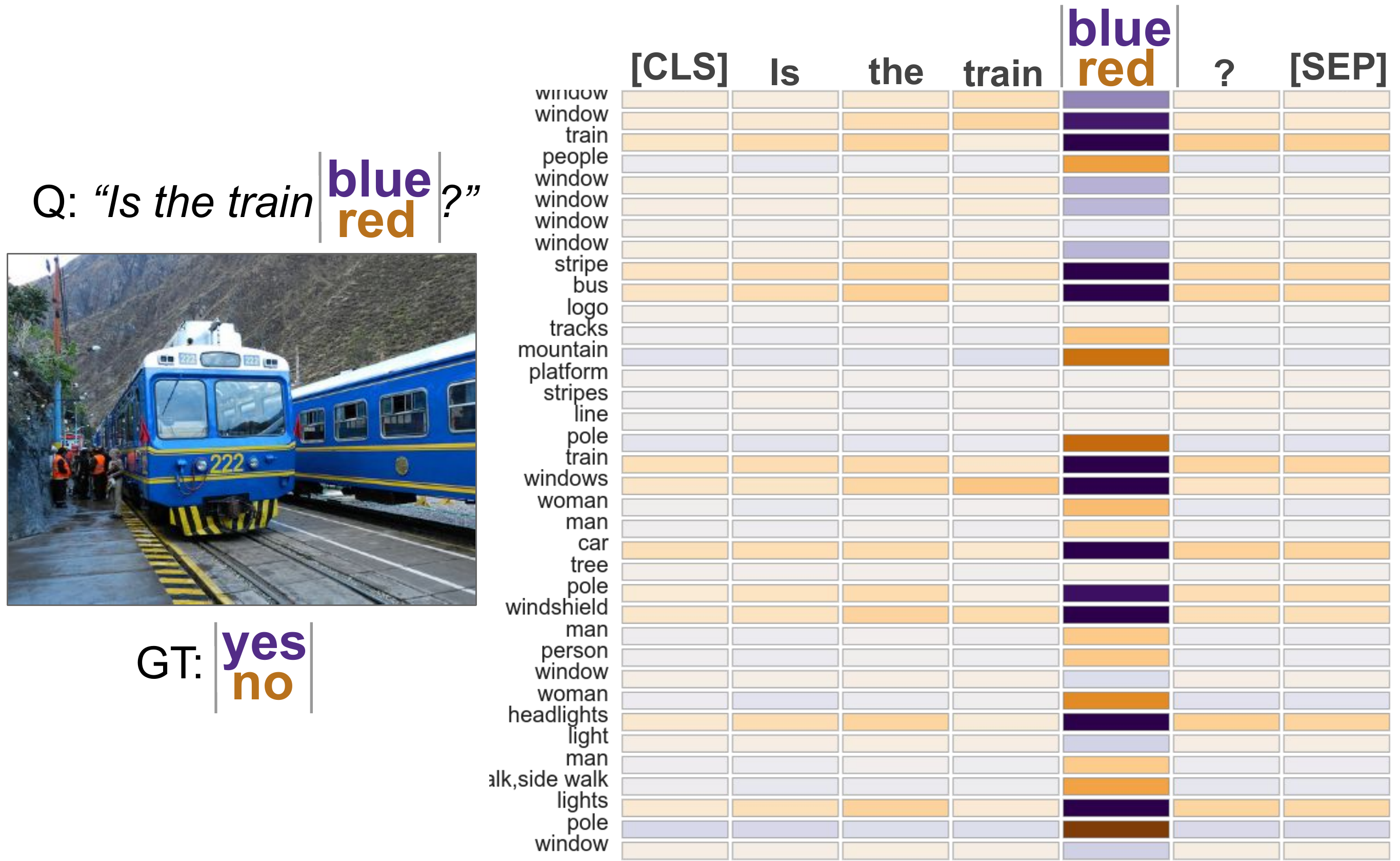}
    \caption{Difference between the attentions of Head LV\_1\_0, when asked ``is the the train blue?'', and ``is the the train red?''. We can observe that in this head, the attention focuses on different objects (row) depending on the color asked (column).}
    \label{fig:headComparison}
        \vspace{-.4cm}
\end{figure}

\myparagraph{Comparison of instances (Fig.~\ref{fig:headComparison})} Comparison between models, and instances can also yield interesting insights on how a model behaves~\cite{derose2020attention}. Currently, to this end, \tool\ memorizes inputs and all intermediate and final results including \emph{k-numbers} and attention maps. This state can be saved, and then used at any time and compared to a new current instance through the \emph{compare} button at the top of the \emph{instance-view}. The comparison itself is obtained by computing the difference of \emph{k-numbers}, and complete attention maps. As a result, head representations in \emph{instance-view} now encode, with a single hue color scale, the difference between the two attention maps. Besides, as shown in Fig.~\ref{fig:headComparison}, attention heat-maps also encodes such a difference using a diverging color scale from dark blue for values close to $-1$, \ie a cell with high attention in the previous instance but not in the current one, to brown for values close to $1$, the opposite. Comparisons are particularly useful when combined with the possibility to ask free-form questions. As it can be observed in fig~\ref{fig:headComparison}, the manually asked questions \emph{``Is the train blue?''} and \emph{``Is the train red?''} are responsible for different attention modes between the word representing the color and bounding boxes of the image. By browsing those bounding boxes, it can be observed that the selected head \emph{LV\_1\_00} associates them, in this case, to the word color only if their color matches. As an example,, in Fig~\ref{fig:headComparison} (right), the third row, which corresponds to the bounding box of a blue train, is associated through attention with the word \emph{``blue''} but not the word \emph{``red''}. In order for such a comparison to be relevant, shifts between two instances must be on a few words of the question, as otherwise, attention maps rows and columns would not align between the instances, and thus any difference would occur for wrong reasons. We plan for future work to enhance this functionality of \tool\ through better visual encoding, more intuitive interactions, and an evaluation.



    

    


\section{Conclusion}

We introduced \tool, an interactive visual analytics tool designed to perform an instance-based in-depth analysis of the reasoning behavior transformer neural networks for vision and language reasoning, in particular visual question answering. \tool\ allows users to select display VQA instances based on the likelihood of bias exploitation; to display attention head intensities; to inspect attention distributions; to prune attention heads; and to directly interact with the model by asking free-form questions. 
Our quantitative evaluations are encouraging, providing first evidence that human users can obtain indications on the reasoning behavior of a neural network using \tool, i.e. estimates on whether it correctly predicts an answer, and whether it exploits biases. \tool\ received positive feedback from these experts, who also provided additional qualitative feedback on the nature of the information they extracted on the behavior of different neural models.


\acknowledgments{ 
We would like to thank reviewers from the previous EuroVis'21 submission, as well as coworkers for their valuable discussions, and proofreading: Nicolas Jacquelin, Aurélien Tabard, and Antoine Coutrot. We finally thank our experts for their time and enthusiasm during the evaluation of \tool. We also acknowledge grant ANR-20-CHIA-0018 ``\emph{Remember}'' of call ``ANR AI chairs \emph{hors centres}''.
}

\bibliography{ref}
\bibliographystyle{template/abbrv-doi}

\end{document}